\documentclass[twoside,11pt]{article}

\usepackage{jmlr2e}
\usepackage{mathtools}
\usepackage{graphicx}
\usepackage{epsfig}
\usepackage{subfigure}

\newcommand{\dataset}{{\cal D}}

\ShortHeadings{A Note on Posterior Probability Estimation for Classifiers}{Nalbantov and Ivanov}
\firstpageno{1}

\begin{document}

\title{A Note on Posterior Probability Estimation for Classifiers}

\author{\name Georgi Nalbantov \email gnalbantov@mdscience.eu \\
	\name Svetoslav Ivanov \email sivanov@mdscience.eu \\
       \addr Department of Data Science, Medical Data Science Ltd., Bulgaria}

\editor{}
\today

\maketitle

\begin{abstract}
One of the central themes in the classification task is the estimation of class posterior probability at a new point $\bf{x}$.
The vast majority of classifiers output a score for $\bf{x}$, which is monotonically related to the posterior probability via
an unknown relationship. There are many attempts in the literature to estimate this latter relationship. Here,
we provide a way to estimate the posterior probability without resorting to using classification scores.
\end{abstract}


\section{Introduction}

The estimation of class posterior probabilities is a central topic in machine learning and statistics.
Most classification algorithms do not model the posterior probability of a given class, say ``$+$'' at point $\bf{x}$ from dataset $\dataset$, $P\left(+ \mid \bf{x} \right)$, 
but output a score that is monotonically related to  $P\left(+ \mid \bf{x} \right)$. Classifiers that do provide (indirect)
estimation of $P\left(+ \mid \bf{x} \right)$ are, for example, logistic regression, linear/quadratic discriminant analysis, and Naive Bayes.
For classifiers that do not provide such estimation, but instead provide a ``score'' for the predicted value for a 
given class at point $\bf{x}$, the practice is to estimate the relationship between score and posterior probability. 
For instance, for support vector machines it is common to use the proposed method by Platt \citep{Platt99probabilisticoutputs} which is based on
negative log-likelihood estimation. More generally, the conformity approach has been proposed that estimates a ``strangeness'' value
from which eventually the posterior probability is derived \citep{vovk2008,vovk2014a}, as well as estimation based on isotonic regression \citep{Zadrozny2002} and Venn-Abers predictions \citep{ayer1955,vovk2014b}. 

The purpose of this paper is to propose a method for the estimation of posterior probability, based on iterative re-building of a given classifier, 
where each time the ratio of negative and positive observations is varied. For shrinkage/penalization methods,
we keep the (effective) total number of points in the dataset fixed, represented by the total sum of weights of the instances. This is a direct approach to the computation of class
membership probability, which does not involve the usage of a score for point $\bf{x}$, which is a common approach.
The paper is organized as follows: Section~\ref{sec:approach} describes the proposed approach to computing the posterior class probability for any classification algorithm; Section~\ref{sec:results}
provides results; Section~\ref{sec:discussion} is devoted to discussion and proposed extensions' and Section~\ref{sec:conclusion} concludes.

\section{A general approach to computing posterior class probabilities}
\label{sec:approach}

Consider the binary classification task of predicting the class of a point $\bf{x}$ given dataset $\dataset: \bf{x}, y \in {\cal R}^n \times \{+,- \} $.
The Bayes rule for computing the posterior probability for class ``$+$'' at point $\bf{x}$ is:

\[
P\left(+ \mid \bf{x} \right) = \frac{P\left( \bf{x} \mid + \right) P\left(+ \mid \bf{x} \right)}{P\left(\bf{x}\right)},
\]

\noindent which in general for continuous $\bf{x}$ is:

\[
\label{eq:condBayes}
P\left(+ \mid \bf{x} \right) = \frac{f\left( \bf{x} \mid + \right) P\left(+ \mid \bf{x} \right)}{f\left(\bf{x}\right)},
\]

\noindent where $f\left( + \mid \bf{x} \right)$ denotes the value of the probability density function for the ``$+$'' class
at point $\bf{x}$. It follows that

\begin{equation}
\frac{P\left(+ \mid \bf{x} \right)}{P\left(- \mid \bf{x} \right)} = \frac{f\left( \bf{x} \mid + \right) P\left( + \right)}{f\left( \bf{x} \mid - \right) P\left( - \right)}.
\end{equation}

In general $P\left( + \right)$ and $P\left( - \right)$, the proportion of ``$+$'' and ``$-$'' points in the population, 
are estimated as the proportion of the ``$+$'' and ``$-$'' points in the training dataset. The estimation
of $f\left( \bf{x} \mid + \right)$ and $f\left( \bf{x} \mid - \right)$ is not straightforward, however.
A central idea we employ is that along the separation surface between the classes we do know that
$P\left(+ \mid \bf{z} \right) = P\left(- \mid \bf{z} \right) = 0.5$, from which it follows that

\begin{equation}
\label{eq:relDensities}
\frac{f\left( \bf{z} \mid + \right)}{f\left( \bf{z} \mid - \right)} = \frac{P\left( - \right)}{P\left( + \right)}
\end{equation}

\noindent for all points $\bf{z}$ along the class separation surface (by definition)\footnote{The logic applies also to implicit classifiers, which do not provide 
a functional form for the separation surface, as the only requirement here is to be able to detect that $P\left(+ \mid \bf{x} \right) = P\left(- \mid \bf{x} \right)$ for a point $\bf{x}$}.
If we alter the (effective) amount of points from the classes so that the new proportions are computed as $P^{new}\left(+ \right)$ and $P^{new}\left(- \right)$, we can recompute for the same $\bf{x}$

\begin{equation}
\label{eq:ff_eq_pp}
\frac{P^{new}\left(+ \mid \bf{x} \right)}{P^{new}\left(- \mid \bf{x} \right)} = \frac{f\left( \bf{x} \mid + \right) P^{new}\left( + \right)}{f\left( \bf{x} \mid - \right) P^{new}\left( - \right)} = 
\frac{P\left( - \right) P^{new}\left( + \right)}{P\left( + \right) P^{new}\left( - \right)},
\end{equation}

\noindent which will not be equal to 0.5. There exist points $\bf{z}$, however, which form the separation surface in the case when the proportions of  and  ``$+$'' and ``$-$''
points in the dataset are $P^{new}\left(+ \right)$ and $P^{new}\left( - \right)$, respectively, for which it holds that

\begin{equation}
\label{eq:pp_z_05}
\frac{P\left(+ \mid \bf{z} \right)}{P\left(- \mid \bf{z} \right)} = \frac{f\left( \bf{z} \mid + \right) P^{new}\left( + \right)}{f\left( \bf{z} \mid - \right) P^{new}\left( - \right)} = 0.5.
\end{equation}

Effectively, we have computed in Eq.(\ref{eq:ff_eq_pp}) the posterior probability at point $\bf{x}$ for a model built on a dataset where the proportions of 
``$+$'' and ``$-$'' points are $P^{new}\left(+ \right)$ and $P^{new}\left( - \right)$, respectively. If we had started the analysis with this dataset at hand (and
refer to it as ``new dataset''), then with respect to the model built on ``new dataset'', which is in practice from now on original dataset we have been provided,
the posterior probability for point $\bf{x}$ (which does not lie on the separation surface with respect to ``new dataset'') is found from Eq.(\ref{eq:ff_eq_pp}) as:

\begin{equation}
\label{eq:pnew_ff_pp}
P^{new}\left(+ \mid \bf{x} \right) = 1/\left({1+  \frac{f\left( \bf{x} \mid + \right) P^{new}\left( + \right)}
{f\left( \bf{x} \mid - \right) P^{new}\left( - \right)}}\right) = 1/\left({1+  \frac{P\left(-\right) P^{new}\left( + \right)}{P\left( + \right) P^{new}\left( - \right)}}\right) ,
\end{equation}

\noindent where $P\left(+\right) $, $P\left(-\right)$, $P^{new}\left(+\right)$, and $P^{new}\left(-\right)$ are estimated as the observed proportions in the data.
The interpretation is as follows:
Given the original data ``new dataset'', and would like to compute $P\left(+ \mid \bf{x} \right)$  with respect to the separation surface
for ``new dataset'', then we have to change the relative weight of ``$+$'' and ``$-$'' points from ``new dataset'' in such a way that
the resulting proportions of ``$+$'' and ``$-$'' points, which approximate $P\left(+\right)$ and $P\left(-\right)$, ensure that the
separation surface pertaining to this ``changed dataset'' (provided by the classification algorithm at hand) goes through point $\bf{x}$. Once this is ensured, we compute 
${f\left( \bf{x} \mid + \right)}/{f\left( \bf{x} \mid - \right)}$ as ${P\left(-\right)}/{P\left(+\right)}$, and plug it in the Bayes formula
for $P^{new}\left(+ \mid \bf{x} \right)$ in Eq.(\ref{eq:pnew_ff_pp}). In general, we do have to perform a search over what values
of $P\left(+\right)$ and $P\left(-\right)$ are needed, but they can be approximated arbitrary well. In case
the classification method in question is a shrinkage/penalization method, like support vectors machines or LASSO, it would be
appropriate to keep the total (effective) number of points fixed, when rebuilding to classification model with varying $P\left(+\right)$ and $P\left(-\right)$, to avoid 
estimation bias. The steps in computing the posterior probability $P^{new}\left(+ \mid \bf{x} \right)$ are summarized in Figure 1 and run as follows:

 \begin{figure*}[h]
	\label{fig:est_steps}
 	\center
  \subfigure[]{
 		\label{fig:est_steps_a}
 		\includegraphics[height=4.196642686cm,width=4.6882494cm]{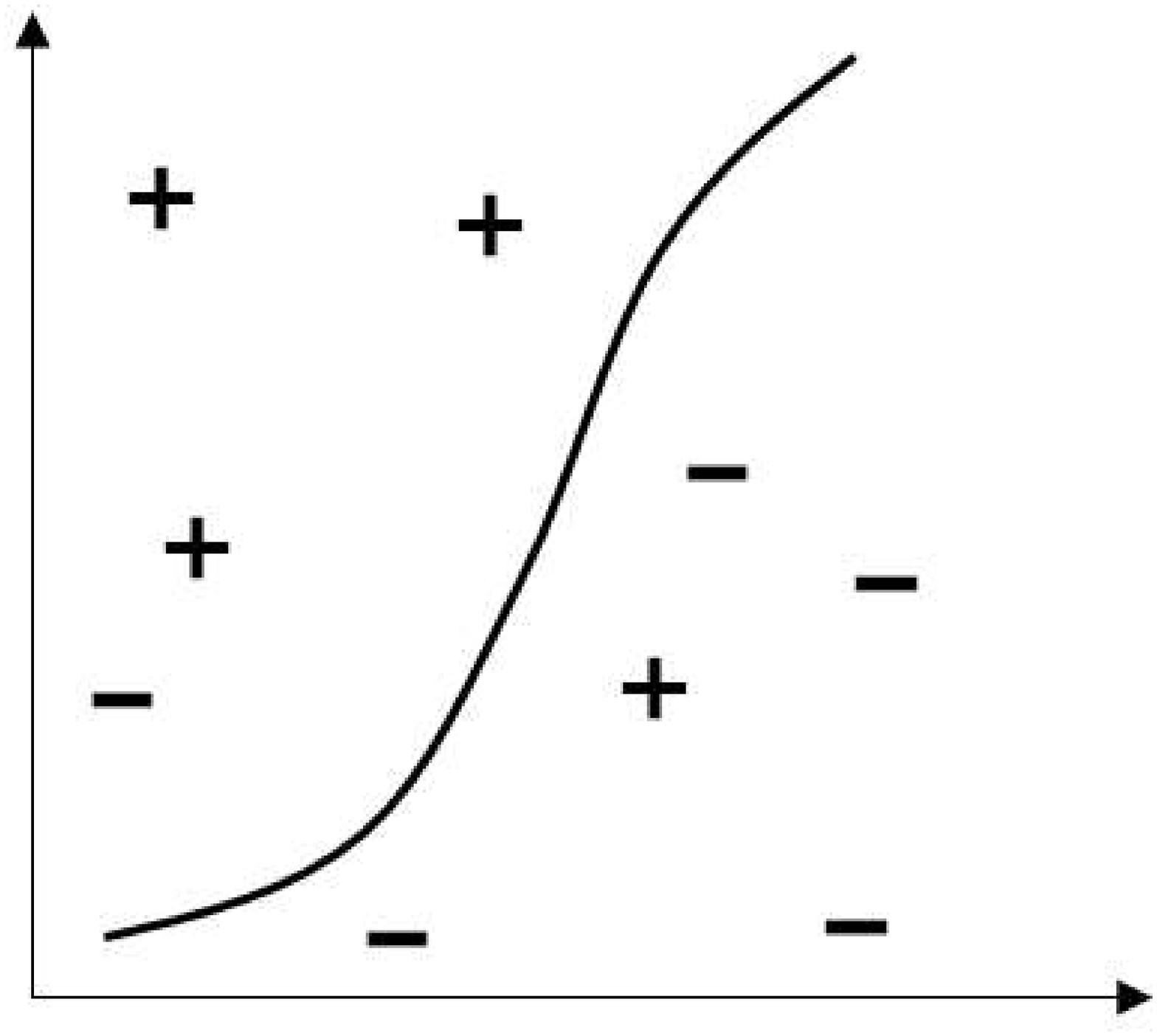}}
 	\subfigure[]{
 		\label{fig:est_steps_b}
 		\includegraphics[height=4.196642686cm,width=4.6882494cm]{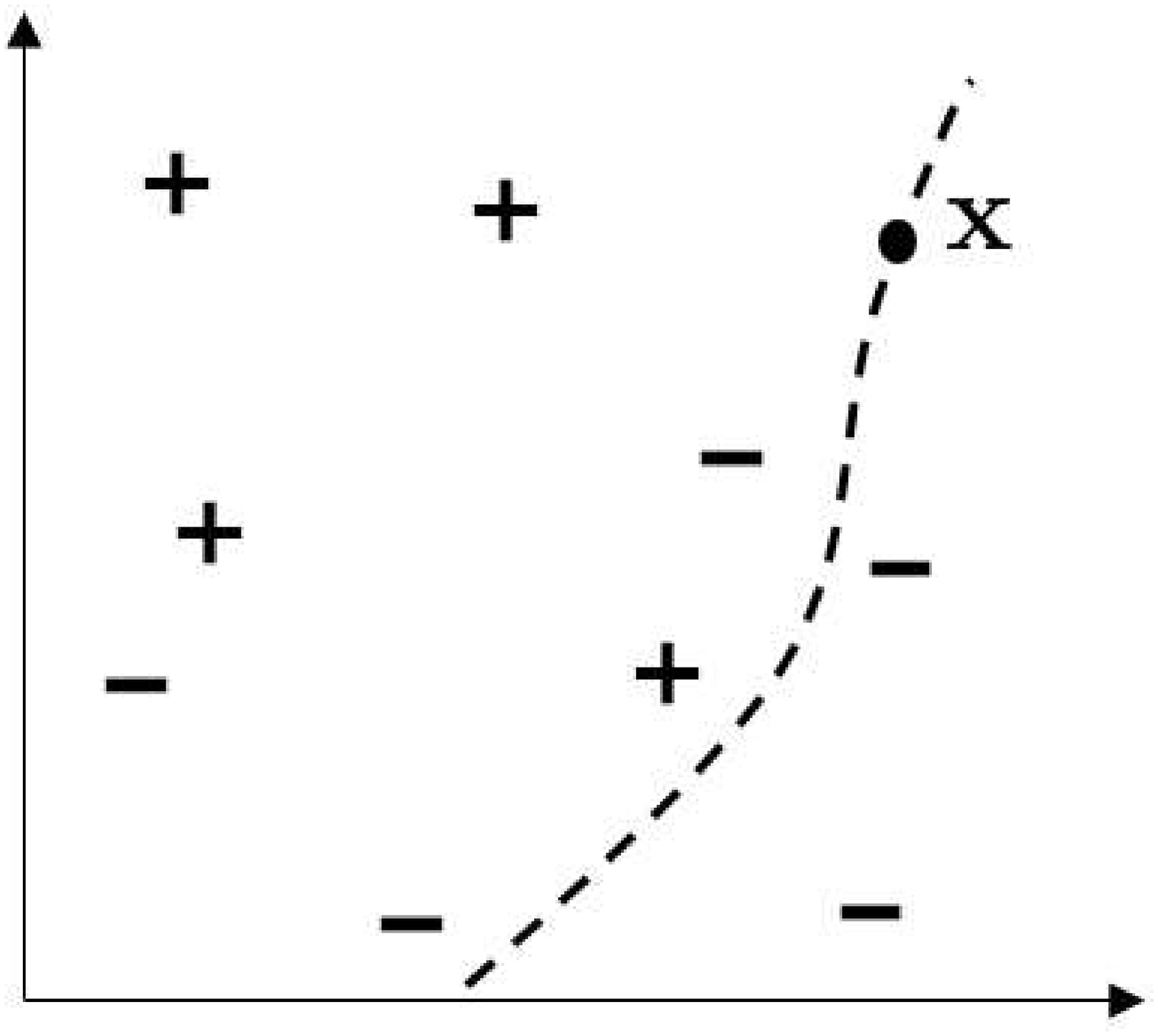}}
 	\subfigure[]{
 		\label{fig:est_steps_c}
 		\includegraphics[height=4.196642686cm,width=4.6882494cm]{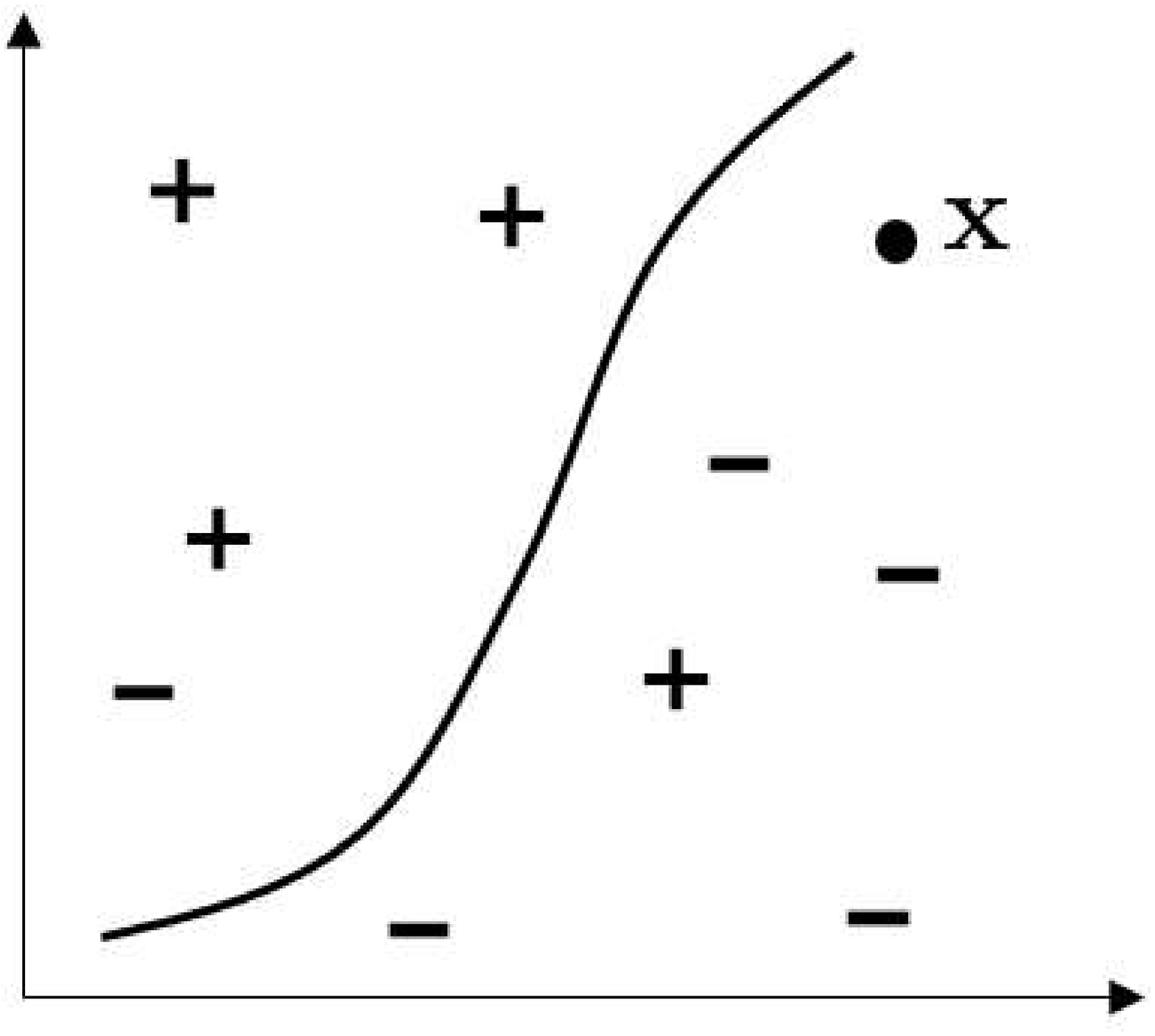}}
 	\caption{Steps for computing posterior probability at point $\bf{x}$ for a given classifier}
 \end{figure*}

Step 1: Figure \ref{fig:est_steps_a}. Consider original training data (referred to as ``new dataset'') with (possibly implicit) separation surface (the solid curve) pertaining to a model built using a given classification algorithm.

Step 2: Figure \ref{fig:est_steps_b}. Change the relative weight of the classes to $P\left(+\right)$ and $P\left(-\right)$ respectively, keeping the effective total number of points constant, until
the re-computed model classifies point $\bf{x}$ as $P\left(+ \mid \bf{x} \right) = 0.5$, implying $ {f\left( \bf{x} \mid + \right)}/{f\left( \bf{x} \mid - \right)}
 =  {P\left( - \right)}/{P\left( + \right)}$.

Step 3: Figure \ref{fig:est_steps_c}. Estimate $P^{new}\left(+ \mid \bf{x} \right)$ for the model built in Step 1 as in Eq.(\ref{eq:pnew_ff_pp}), where $P\left(+\right) $, $P\left(-\right)$, $P^{new}\left(+\right)$, and $P^{new}\left(-\right)$ are estimated directly from the data as respective proportions.

\section{Experimental results on a 2D toy dataset}
\label{sec:results}

We illustrate the approach for estimating posterior probabilities for classifiers on a 2D two-class toy dataset. Each class in the dataset was generated by drawing 1000 random samples from a Gaussian distribution with different means and the same covariance matrix. The classifiers used in this example are linear support vector machines (SVM), logistic regression, and decision trees.

The linear SVM classifier's $C$ parameter was fixed at 1. After the SVM was run, all points which are not support vectors were removed. In this way we ensure that the estimation of iso-probability curves is carried out on the points (support vectors), which were used to create the decision surface. The resulting iso-probability curves for levels 0.05 - 0.95 (with step 0.05) are plotted in Figure 2. The total number of points (represented as the total sum of weights) was kept fixed in the computation of these curves. We note that different values for the $C$ parameter will produce different sets of iso-probability curves, as this parameter influences the flatness of the SVM separation surface. Thus, each $C$ parameter effectively produces a different classifier, which has an intrinsic probability estimate for a given point.

 \begin{figure*}[h]
 	\center
 	\hspace*{-2cm}
 		\label{fig:svm_dataset}
 		\includegraphics[angle=-90,origin=c,height=13cm]{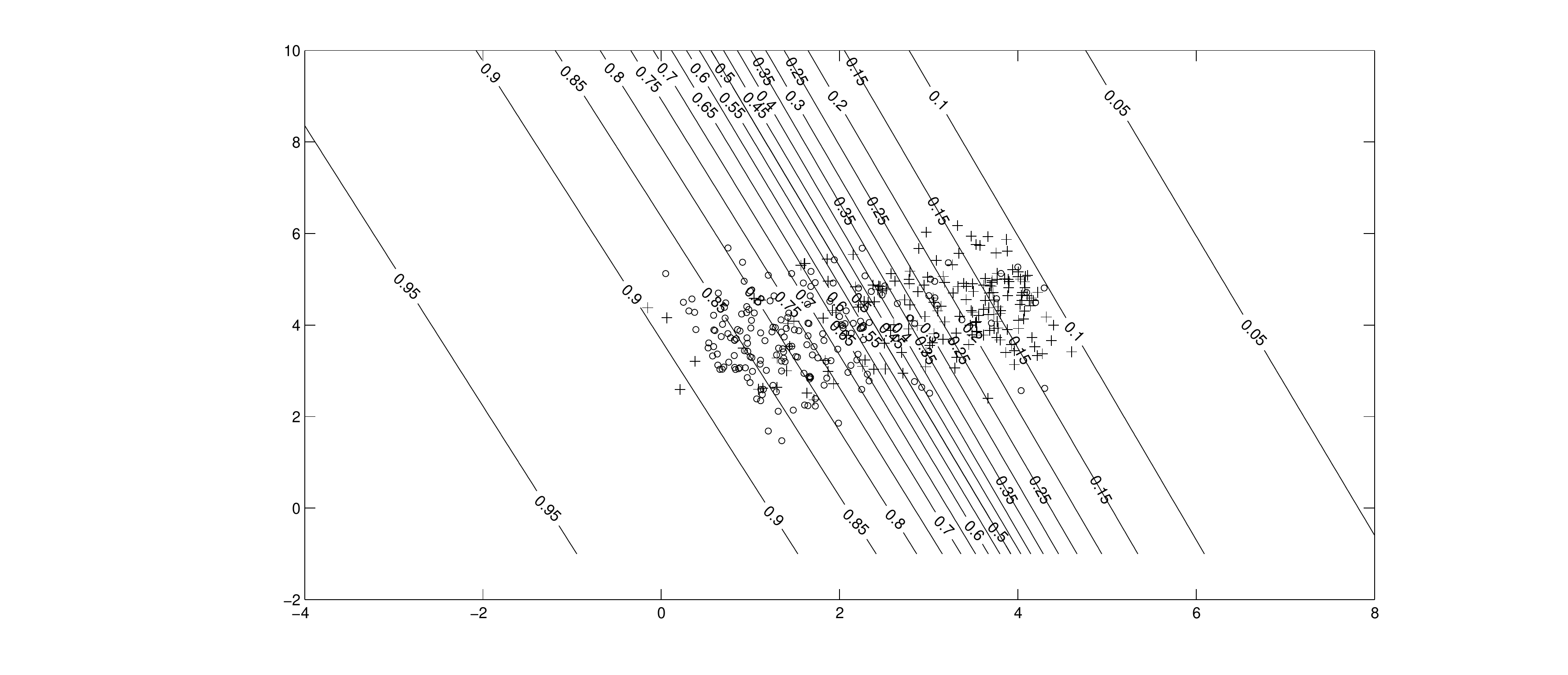}
 	\vspace*{-6cm}
	\caption{Iso-probability curves for SVM}
 \end{figure*}

\begin{figure*}[h]
 	\center
 	\hspace*{-0.85cm}
 		\label{fig:logit_dataset}
 		\includegraphics[angle=-90,origin=c,height=13cm]{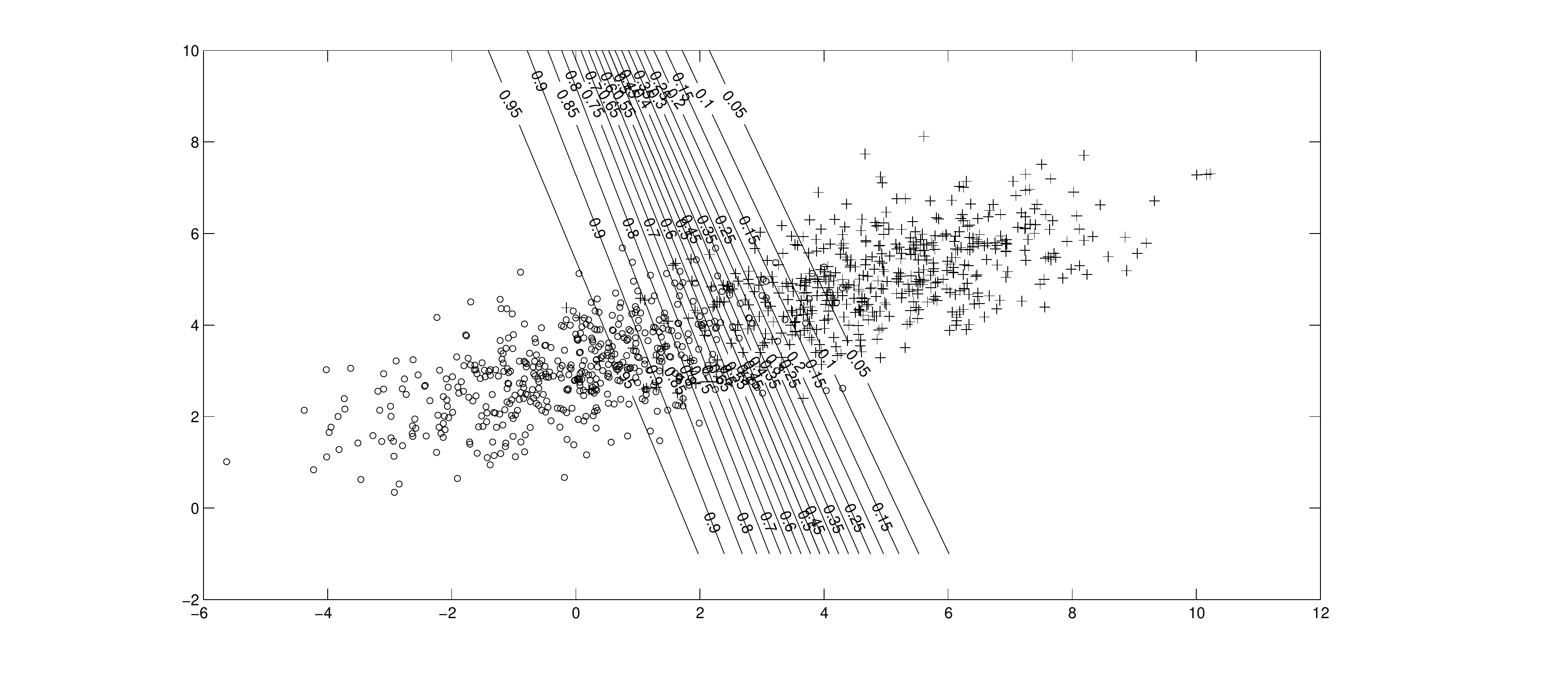}
 	\vspace*{-6cm}
 	\caption{Iso-probability curves for logistic regression}
 \end{figure*}

The results for the (linear) logistic regression classifier are shown in Figure \ref{fig:logit_dataset}. The iso-probability curves for levels 0.05 - 0.95 (with step 0.05) are plotted.

\begin{figure*}[h]
 	\hspace*{-0.85cm}
 		\label{fig:dt_dataset}
 		\includegraphics[height=13cm]{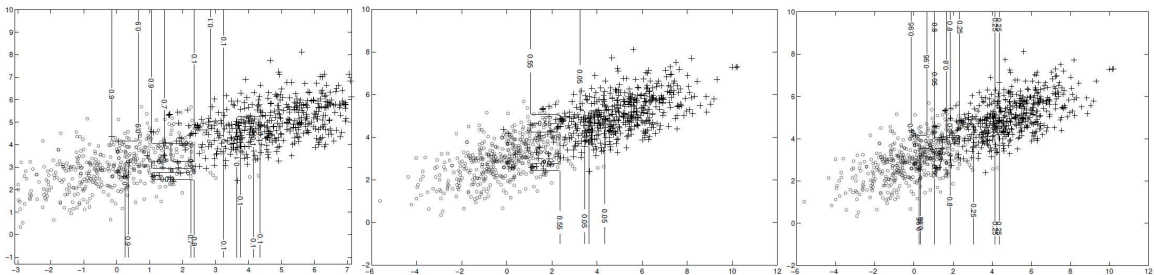}
 	\vspace*{-9cm}
 	\caption{Iso-probability curves for decision trees}
 \end{figure*}

The results for the decision tree classifier are shown in Figure \ref{fig:dt_dataset}. Prunning was applied. The iso-probability surfaces for selected levels (from left to right subfigures: 0.1, 0.7, 0.9, then 0.05, 0.55, and finally 0.25, 0.8, 0.95) are plotted. Notably, the iso-probability surfaces coinside partially at different iso levels, which might be a general property linked to the decision tree estimation process.

For SVM and logistic regression we present the relationship between (raw) scores and estimated class posterior probabilities in Figure 5. The resolution is 0.05 as this is the step used for computing the iso-probability curves. For decision trees it is not possible to present a similar figure, as the classifier does not output prediction scores, but rather predicted class labels.


\section{Discussion}
\label{sec:discussion}


We have proposed a general, classifier-independent method for computing posterior class probability which only requires that
the classifier can detect whether a point in question belongs to the separation surface or not (which is trivial for any classifier)
and that the estimates of population proportions are computed as the observed proportions in the training data.
We have addressed explicitly the binary classification problem. Because the Bayes formula holds for any class in a multi-class problem,
the extension in this respect is straightforward. One choice that was made was to change the relative proportion of the two classes
in the data by changing the ``weights'' of the points from each class (equally for points belonging to the same class), while keeping the total effective number of points fixed, rather than removing
points/observations randomly from the classes, for reason of robustness of estimation. In a number of classifiers changing the weight of points
is trivial and can be done directly in the classifier's estimation/optimization (where applicable) procedure. For example, for support vector machines
it is sufficient to change the so-called $C^+$ and $C^-$ parameters in order to obtain new effective number of points in a dataset.

We note that there is no explicit guarantee that the iso-probability curves produced by a classifier on a given dataset are not crossing each other, even if the classifier is linear. 

 \begin{figure*}[h]
 	\center
	\hspace*{-0.9cm}
 	\hspace*{0cm}
 		\label{fig:calib_SVM_Logit}
 		\includegraphics[height=24cm]{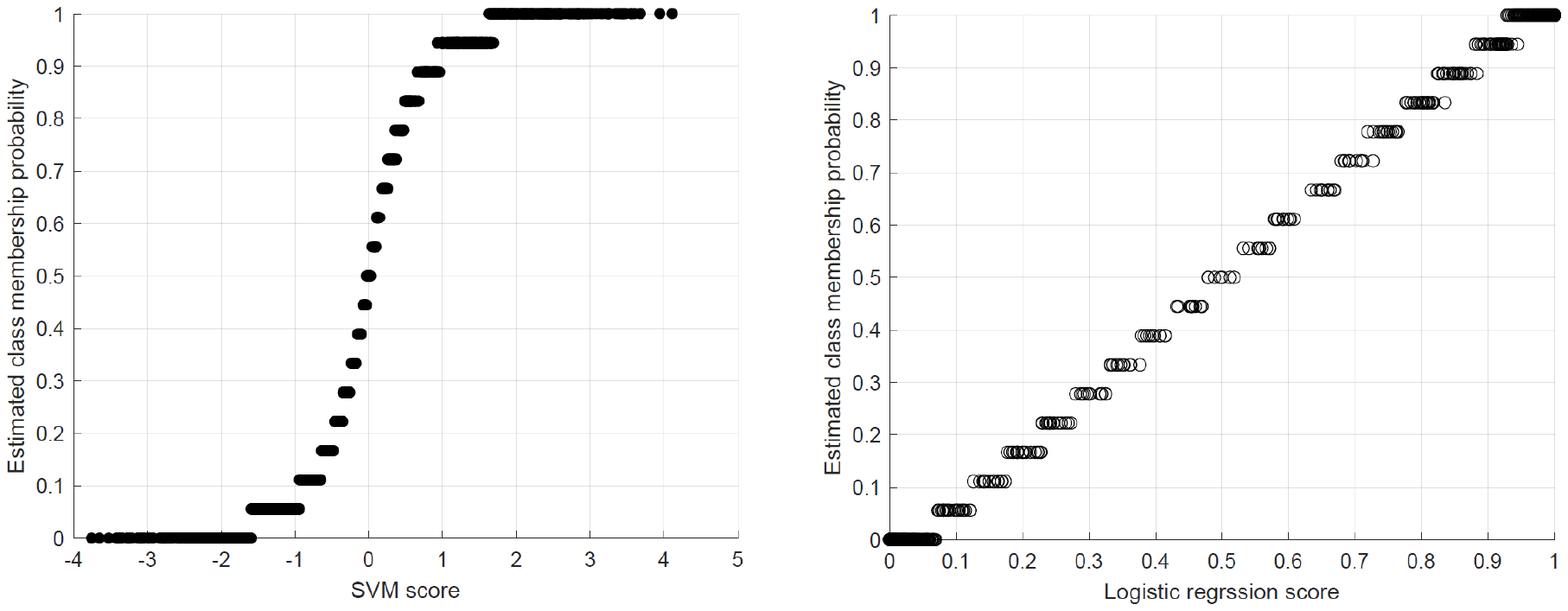}
 	\vspace*{-17.2cm}
 	\caption{Estimated raw scores vs. estimated class membership probability for SVM (left) and logistic regression (right)}
 \end{figure*}

This degeneracy should be exploited further, as it suggests that one datapoint can be associated with two different class posterior probability estimates. This degeneracy is illustrated
in Figure 6. A classfication algorithm may produce two separation surfaces that cross each other: the first one is built using weights on the 
positive and negative instances equal to 1.5, while second one is built using corresponding weights of 1 and 2. At the intersection point the probability of each class is equal to 0.5 in both
cases, which leads to two different calculations  from Eq.(\ref{eq:relDensities}) of the relative class densities at the intersection point. Clearly, this ``estimation degeneracy''
problem would tend to disappear for classifiers which are consistent with the Bayes rule as the number of points in the data goes large. A possible approach to deal with this degeneracy would be to impose monotonicity and apply isotonic regression for calibrating the probability estimates. Another candidate approach is to consider the estimated different posterior probabilities as upper and lower bounds of the estimation procedure.

In this note we have abstained from comparison with alternative methods for estimation of posterior probability due to the inability to provide
a comprehensive comparison in just one manuscript, as the alternative approaches are numerous.

\section{Conclusion}
\label{sec:conclusion}

 \begin{figure*}[t]
 	\center
 		\label{fig:fig_deg}
 		\includegraphics[height=5cm]{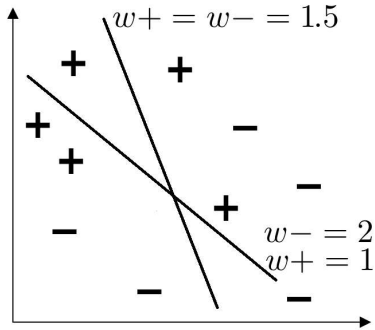}
 	\vspace*{0cm}
 	\caption{Illustraction of a degeneracy. A classfication algorithm produces two separation surfaces that cross each other when the relative weights of the positive $w+$
and negative $w-$ instances are changed.}
 \end{figure*}

The usual approach to solving the task of finding the class posterior probability 
is to choose a test point $\bf{x}$ for which to estimate this probability. In contrast, 
we have approached the same task from a different angle: rather than focusing on $\bf{x}$,
focus instead on finding iso-surfaces with constant posterior probabilities pertaining to a given classification
model. The iso-surfaces are obtained by varying the relative proportion (weight) of observations belonging to different classes.
This approach thus avoids any reference to observations' ``scores'' from which to estimate further on posterior probabilities.
We envisage further explicit extension of the approach to the multi-class classification task, as well as a possible 
extension to the regression task.

\vskip 0.2in
\bibliography{NoteOnPosteriorClassProbablity}

\end{document}